\documentclass[final]{cvpr}

\usepackage{times}
\usepackage{epsfig}
\usepackage{graphicx}
\usepackage{amsmath}
\usepackage{amssymb}
\usepackage{algorithm}
\usepackage{algorithmic}

\usepackage[pagebackref=true,breaklinks=true,colorlinks,bookmarks=false]{hyperref}

\def\cvprPaperID{11578} 
\def\confYear{CVPR 2021}

\begin{document}

\title{Brain Image Synthesis with Unsupervised Multivariate Canonical CSC$\ell_4$Net}

\author{Yawen Huang\textsuperscript{1}, Feng Zheng\textsuperscript{2}\thanks{Corresponding author}, Danyang Wang\textsuperscript{1}, Weilin Huang\textsuperscript{1}, Matthew R. Scott\textsuperscript{1}, Ling Shao\textsuperscript{3}\\
\and
\textsuperscript{1}Malong LLC, \textsuperscript{2}Southern University of Science and Technology, \textsuperscript{3}Inception Institute of Artificial Intelligence\\
{\tt\small \{yawhuang, danwang, whuang, mscott\}@malongtech.com, zhengf@sustech.edu.cn, ling.shao@ieee.org}}

\maketitle

\begin{abstract}
	Recent advances in neuroscience have highlighted the effectiveness of multi-modal medical data for investigating certain pathologies and understanding human cognition. However, obtaining full sets of different modalities is limited by various factors, such as long acquisition times, high examination costs and artifact suppression. In addition, the complexity, high dimensionality and heterogeneity of neuroimaging data remains another key challenge in leveraging existing randomized scans effectively, as data of the same modality is often measured differently by different machines. There is a clear need to go beyond the traditional imaging-dependent process and synthesize anatomically specific target-modality data from a source input. In this paper, we propose to learn dedicated features that cross both intre- and intra-modal variations using a novel CSC$\ell_4$Net. Through an initial unification of intra-modal data in the feature maps and multivariate canonical adaptation, CSC$\ell_4$Net facilitates feature-level mutual transformation. The positive definite Riemannian manifold-penalized data fidelity term further enables CSC$\ell_4$Net to reconstruct missing measurements according to transformed features. Finally, the maximization $\ell_4$-norm boils down to a computationally efficient optimization problem. Extensive experiments validate the ability and robustness of our CSC$\ell_4$Net compared to the state-of-the-art methods on multiple datasets.
\end{abstract}
\begin{figure}[!t]
	\centering
	\includegraphics[width=1\linewidth]{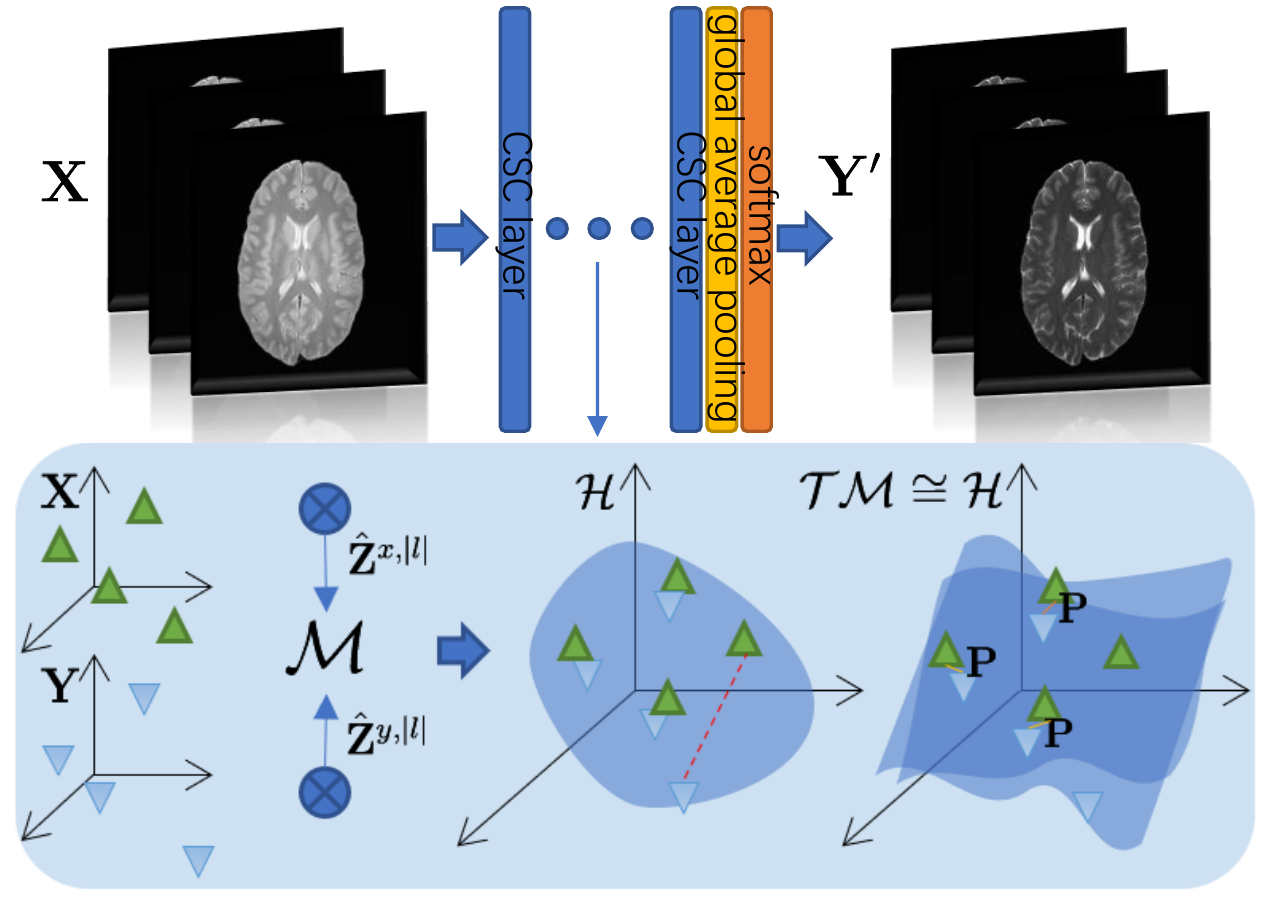}
	\caption{Architecture of our CSC$\ell_4$Net. CSC$\ell_4$Net is constructed by repeatedly stacking multivariate canonical CSC layers. The blue bar denotes the CSC layer, the yellow bar is the global average pooling, and the orange bar shows the softmax. $\hat{\mathbf{Z}}^{x,\left | l \right |}$ and $\hat{\mathbf{Z}}^{y,\left | l \right |}$ are feature maps of the $l$-th layer. $\mathcal{M}$ represents the manifold on each tangent space $\mathcal{T}$ having $\mathcal{T}\mathcal{M}$. $\mathbf{P}$ is the associator and $\mathcal{H}$ denotes a Hilbert space.}
	\label{fig5}
	\vspace{-0.42cm}
\end{figure}

\section{Introduction}
Craniocerebral examination can be carried out using a multitude of imaging techniques with varying degrees of specificity and invasiveness, each directly or indirectly quantifying the structure, function and pathology of the brain. These multi-modal neuroimaging techniques, such as magnetic resonance imaging (MRI) and positron emission tomography (PET), offer diverse and complementary information to investigate human cognitive activities, population imaging cohorts, neurodegeneration, and certain pathology. However, acquiring a full library of multi-modal images is impractical since the collection faces several constraints, including long acquisition times (\textit{e.g.} a normal MRI scan can take as long as an hour), high examination cost, or even worse, image corruption in the event of artifacts from patient motion. Missing data is a critical problem in neurological studies and clinical diagnosis~\cite{huang2017cross}, and thus there is a clear need to obtain the absent data through beyond simple scanning.

Recently, there has been a surge of interest in synthesizing target-modality medical images by transferring information across different appearances~\cite{huang2017dote,vemulapalli2015unsupervised,wang2018high,zhou2019high}. One early and noteworthy model for this was joint sparse representation~\cite{huang2017cross,roy2013magnetic}, which allows multi-modal data to be mapped in a common space rather than in separate ways to obtain a linear approximation. Based on this, the convolutional sparse coding (CSC) model replaces the local optimization with a global shift-invariant one, achieving significant improvements~\cite{gu2015convolutional,heide2015fast,sun2019adversarial}. Further, deep learning has obtained promising results in multi-modal image synthesis, mostly with convolutional neural networks (CNNs)~\cite{gatys2016image} and generative adversarial networks (GANs)~\cite{zhang2018translating,zhou2019high}.

While synthesis methods have had significant impact on research, there is now a debate regarding whether such synthetic images can substitute real acquisitions in clinical analyses. In general, clinical diagnosis requires multiple biomarkers to identify the disease and its status. When the target acquisitions are missing, accurate synthesis is essential. Although this challenge has been tackled by generating objective modality byproducts, current results remain unacceptable for clinical diagnosis. 

In addition to differences in modalities, the complexity, high-dimensionality and heterogeneity of medical data remains another key challenge in leveraging existing randomized scans effectively. Specifically, imaging using machines developed by different manufacturers (\textit{e.g.} Philips, Siemens, GE, \textit{etc.}), the abundance of various physical parameters, and the presence of temporal dependency, all introduce conflicted and inconsistent features, thus preventing the complete use of real acquired data. It is nontrivial to harmonize all the different information and construct their correlations, but efforts to address the above challenges and develop a reliable algorithm to effectively utilize data are extremely necessary for both research and clinical decision support.

In this paper, we propose an unsupervised multivariate canonical CSC$\ell_4$Net, a novel approach to crossing both intra-modal (\textit{i.e.} more than one measurement from the same data modality) and inter-modal (\textit{i.e.} more than one data modality) heterogeneities. Our model synthesizes anatomically specific target modality data from a source modality, and makes efficient use of real acquisitions. CSC$\ell_4$Net works well under multiple datasets, despite the high dimensionality, temporal dependency and irregularity of neuroimaging, making it possible to combine acquisitions from different scanner manufacturers by initially normalizing differences between features, and then mapping them into the Hilbert space for multivariate canonical adaptation. Both cross-modal geometry transformations and a neuroimaging-specific positive definite conditions are incorporated within a Riemannian manifold. Finally, solving an $\ell_4$-maximization instead of an $\ell_1$-minimization problem enables us to employ the lowest sample complexity for high computational medical data. An overview of our CSC$\ell_4$Net is shown in Fig. \ref{fig5}.

To summarize, this paper provides the following contributions:
\begin{itemize}
	\setlength{\itemsep}{0pt}
	\setlength{\parsep}{0pt}
	\setlength{\parskip}{0pt}
	\item To the best of our knowledge, this is the first work to generate anatomically meaningful images, by modeling an unsupervised multivariate canonical CSC$\ell_4$Net.	
	\item We propose a novel intra-modal unit normalization for the initial unification of variate data of the same modality to guarantee a unique convolutional sparse solution.
	\item The multivariate canonical feature mapping is formulated over the multi-layer CSC to optimize the inter-modal structure.
	\item We introduce a Riemannian manifold-penalized transformation data fidelity term under the positive definite condition, for which we show how a reformulation based on CSC is crucial to empirical success.
	\item We prove that maximizing the $\ell_4$-norm instead of minimizing the $\ell_1$-norm leads to the lowest complexity and the highest robustness. 
\end{itemize}

\section{Related Work}
\subsection{Image Synthesis}
Image synthesis (\textit{a.k.a.} image-to-image translation) is commonly performed via appearance transformations, such as linear regression or distribution transformation. Previous works with stand-alone image pairs tend to focus on constructing a linear relationship in different contrasts~\cite{roy2013magnetic}. Limited by few off-the-shelf paired data, Vemulapalli~\textit{et al.}~\cite{vemulapalli2015unsupervised} relaxed the invariable supervision by matching similarities across different modalities of image patches, and then jointly maximizing both global mutual information and local spatial consistency. Huang~\textit{et al.}~\cite{huang2017cross} proposed a data-efficient synthesis method by mapping both a few pairs and large amounts of unpaired patches into a high-dimensional space, and then adopting Laplacian eigenmaps for geometric co-regularization. Neural style transfer~\cite{gatys2016image} is another popular strategy for content-fixed image style translation, which computes the distance via the Gram matrix statistics of pre-trained deep features. GAN~\cite{goodfellow2014generative} was introduced to generate images using a random noise vector with discriminator judgment. Subsequently, many improvements and task-oriented generative models have been proposed. CycleGAN~\cite{zhu2017unpaired} uses a simple yet efficient strategy with cycle-consistent adversarial networks for unsupervised image-to-image translation. To synthesize high-resolution images from semantic labels, Wang~\textit{et al.}~\cite{wang2018high} proposed an adversarial learning objective which leverages GANs in a conditional setting and discards hand-crafted losses or pre-trained networks. FUNIT~\cite{liu2019few} contains a content and class-leveled encoder and an overall decoder for synthesizing the analogous image in a desirable category from the given class input. Leveraging the U-Net architecture, a data augmentation method~\cite{zhao2019data} was established by learning both spatial deformation fields and intensity transforms to generate samples. TrGAN~\cite{wang2020transformation} focus on improving unsupervised image synthesis and representation learning.

\subsection{Medical Image Analysis}
The existing studies on medical imaging~\cite{huang2017dote,zhou2019high,huang2020mcmt}, \textit{e.g.} synthesis, segmentation and registration, have shown great promise for either research purposes or clinical analysis, mostly toward a macro objective, \textit{i.e.} computer-aided diagnosis (CAD). In~\cite{iglesias2018joint}, a probabilistic model was proposed for joint registration and synthesis with cross-modality alignment. Uzunova~\textit{et al.}~\cite{uzunova2019multi} used a multi-scale GAN to generate large amounts of high-quality medical images by learning a growing resolution conditioned on front scales. Shao~\textit{et al.}~\cite{shao2019diagnosis} presented a diagnosis-guided multi-modal feature selection method for prognostic prediction of a specific disease. Ravi~\textit{et al.}~\cite{ravi2019degenerative} employed adversarial learning in their proposed DaniNet, a degenerative adversarial neuroimage network that allows neurodegeneration to be modeled.

One of the fundamental purposes of medical image processing is to implement CAD, which in turn allows clinicians to make accurate decisions or provide treatment. Toward generalization and practicability, our work is complementary to the aforementioned approaches, where we improve the usage of large-scale heterogeneous medical data in an unsupervised manner.

\subsection{Convolutional Sparse Coding}
Convolutional sparse coding (CSC) has been successfully used in a wide range of computer vision and medical image processing problems~\cite{bristow2013fast,gu2015convolutional,heide2015fast}. CSC deals with the suboptimality of conventional local-independent representations by introducing a global shift-invariant filter. Fast CSC model was proposed by Bristow~\textit{et al.}~\cite{bristow2013fast}, where the quad-decomposition solves the CSC objective in the Fourier domain, resulting in fast training. CSC-SR, introduced by Gu~\textit{et al.}~\cite{gu2015convolutional},  uses CSC with improved consistency to super-resolve natural images. Heide~\cite{heide2015fast} tackled feature learning with fast and flexible CSC. Huang~\textit{et al.}~\cite{huang2017dote} constructed two invertible mappings based on CSC for cross-modality synthesis and super-resolution of brain images. To reconstruct clean images, while avoiding adversarial attacks, Sun~\textit{et al.}~\cite{sun2019adversarial} presented a stratified CSC algorithm with the benefit of an input transformation-based defense.

\section{Mathematical Description of CSC}
Given samples $\{\mathbf{x}_1,\mathbf{x}_2,...\mathbf{x}_S\}$ in $\mathbb{R}^N$, the problem of learning a set of convolutions of sparse feature maps $\mathbf{z}_{k} \in \mathbb{R}^N$ by filters $\mathbf{f}_{k} \in \mathbb{R}^M$, $\forall k=\left \{ 1,...,K \right \}$ can be expressed as minimizing the optimization function that combines the least-squares error and the $\ell_1$-norm penalty on the representations:
\begin{equation}
\begin{aligned}
\label{eq1}
\arg \min_{\mathbf{f},\mathbf{z}} \frac{1}{2}\left \| \mathbf{x}-\sum_{k=1}^{K}\mathbf{f}_{k}\ast \mathbf{z}_{k} \right \|_{2}^{2}+\lambda \sum_{k=1}^{K}\left \| \mathbf{z}_{k} \right \|_{1}\\
s.t. \, \left \| \mathbf{f}_{k} \right \|_{2}^{2}\leq 1 \; \forall k=\left \{ 1,...,K \right \},
\end{aligned}
\end{equation}
where $\ast$ represents the 2D convolution operator and $\lambda$ denotes a regularization parameter for the sparsity of the $\ell_1$-norm. The objective of Eq. (\ref{eq1}) is not jointly convex with respect to $\mathbf{f}$ and $\mathbf{z}$, but is convex in optimizing one variable while fixing the other~\cite{zeiler2010deconvolutional}. On the basis of such a convex optimization theory, the alternating direction method of multipliers (ADMM) was presented to solve the augmented Lagrangian formulation by introducing more proxies solving in the Fourier domain. ADMM transforms the convolution operation to an element-wise multiplication in order to speed up the spatial dominated convolution, \textit{i.e.}, from $\mathcal{O}(KNM)$ time complexity in the spatial space to $\mathcal{O}(KN\log N)$ in the frequency domain. Despite the accelerated computation attained by FCSC~\cite{heide2015fast}, the auxiliary variables are still deemed heavy training parts, especially for tuning. More recent works~\cite{moreau2018dicod,zisselman2019local} alleviate such a limitation by forming the coordinate descent solution in a local greedy fashion.

Once the biconvex problem of Eq. (\ref{eq1}) has been solved by alternating between learning filters and learning feature maps, the reconstruction can be obtained by a summation of the convolution outputs (\textit{i.e.} convolving every row of $\mathbf{f}$ by $\mathbf{z}$), leading to ${\mathbf{x}}'=\mathbf{f}_{k} \ast \mathbf{z}_{k}$.

\section{Multivariate Canonical CSC$\ell_4$Net}
\subsection{Intra-Modal Unit Normalization}
Let $\mathbf{X} = \{\mathbf{x}_1, ..., \mathbf{x}_S\}$ be a source domain training set containing $S$ source modality images, and $\mathbf{Y} = \{\mathbf{y}_1, ..., \mathbf{y}_T\}$ be a target domain training set containing $T$ target modality examples. We define $\mathbf{F}$ as the filter of the convolutional feature maps $\mathbf{Z}$. When standard training following the independence assumption illustrated in Eq. (\ref{eq1}) is applied to two modalities, this leads to two separate filters of shift-invariant atoms $\mathbf{F}^x$, $\mathbf{F}^y$, with their corresponding feature maps $\mathbf{Z}^x$ and $\mathbf{Z}^y$. 
However, when the same modality data are sampled from random measurements (\textit{i.e.} intra-modal variation), the new features of these variables are no longer a CSC solution. To overcome this problem, we attempt to regularize the intra-modal data to guarantee a unique convolutionally sparse solution. Considering that a unit normalization of data is taken from the random measurements of one modality, one possibility is to normalize the column elements of $\mathbf{Z}^x$ and $\mathbf{Z}^y$ to unity, \textit{i.e., $\left \| \mathbf{Z}^x_i \right \|^2_2 = 1$, $\left \| \mathbf{Z}^y_j \right \|^2_2 = 1$, $\forall i = 1,...,S$, $\forall j = 1,...,T$}. However, using the unit normalization leads to "erased" modality information, which is especially the case in the context of neuroimaging data. To circumvent this issue, we eliminate the intra-modal scaling ambiguity by first computing the maximum $\ell_2$ norm of $\mathbf{Z}^x$ and $\mathbf{Z}^y$, respectively, and then performing our intra-modal unit normalization (IUN). The IUN (termed as $\Upsilon$) of the convolutional feature maps becomes
\allowdisplaybreaks[4]
\begin{align}
\label{eq1.1}
&\hat{\mathbf{Z}}^x_i=\frac{\mathbf{Z}^x_i}{\max (\left \| \mathbf{Z}^x_i \right \|_2)\sqrt{1-\left \| \mathbf{Z}^x_i \right \|^2}},\forall i = 1,...,S, \notag \\ &\hat{\mathbf{Z}}^y_j=\frac{\mathbf{Z}^y_j}{\max (\left \| \mathbf{Z}^y_j \right \|_2)\sqrt{1-\left \| \mathbf{Z}^y_j \right \|^2}},\forall j = 1,...,T.
\end{align}

The restricted elements of $\mathbf{Z}^x$ and $\mathbf{Z}^y$ need to be unified by IUN and satisfy the general unit normalization $\left \| \hat{\mathbf{Z}}^x_i \right \|^2_2 = 1$, $\forall i$, and $\left \| \hat{\mathbf{Z}}^y_j \right \|^2_2 = 1$, $\forall j$.

\subsection{Multivariate Canonical Feature Mapping}
The CSC model of Eq. (\ref{eq1}) has the advantage of computational efficiency when compared to the network-relevant methods. However, the pure penalty under $\ell_1$-norm ignores the diversity and complexity of data, leading to unsatisfactory results. Recent studies~\cite{goodfellow2014generative,feng2019Pyramidal} have shown that, by stacking multiple layers on top of each other, the extracted features are deeper and thus the performance of applications (\textit{e.g.} classification or reconstruction) can be further boosted. Inspired by the success of extensive efforts on networks, we construct a CSC network architecture with a novel multivariate canonical adapted feature mapping layer, which has cross-modal learning power with synthetic potential. 

Without loss of generality, we consider the CSC over IUN, which has multiple layers (termed as CSC-Net). We denote a function $f(\cdot,\cdot)$ for the feature mapping of each layer, such that ${\mathbf{Z}^x}'=f(\mathbf{X}, \mathbf{\Psi}^x)$, ${\mathbf{Z}^y}'=f(\mathbf{Y}, \mathbf{\Psi}^y)$, where $\mathbf{\Psi}^x=\{\mathbf{F}^x, \Upsilon, \lambda\}$, $\mathbf{\Psi}^y=\{\mathbf{F}^y, \Upsilon, \lambda\}$. Let $l$ be the network layers, where $l \in [1, L]$. In CSC-Net, the feature maps $\hat{\mathbf{Z}}$ can be defined as the representation of the layer $l$ with tensor properties of height $h$ and width $w$: $\hat{\mathbf{Z}}^{x,\left | l \right |}_i \in \mathbb{R}^{N^{\left | l \right |} \times h^{\left | l \right |} \times w^{\left | l \right |}}$, $\forall i,l$, and $\hat{\mathbf{Z}}^{y,\left | l \right |}_j \in \mathbb{R}^{N^{\left | l \right |} \times h^{\left | l \right |} \times w^{\left | l \right |}}$, $\forall j,l$. To hierarchically approximate the convolutional feature maps, we construct a sparse intermediate representation which imposes the same structure on the upper layer of the representation. Intuitively, the $l$-th layer of $\mathbf{Z}$ for $\mathbf{X}$ and $\mathbf{Y}$ can be estimated as $\hat{\mathbf{Z}}^{x,\left | l \right |}_i =f(\hat{\mathbf{Z}}^{x,\left | l-1 \right |}_i, \mathbf{\Psi}^{x,\left | l-1 \right |})$ and $\hat{\mathbf{Z}}^{y,\left | l \right |}_j =f(\hat{\mathbf{Z}}^{y,\left | l-1 \right |}_j, \mathbf{\Psi}^{y,\left | l-1 \right |})$, respectively. 

In addition to the multi-layer sparse and deeper representation, another core module of the CSC-Net is the multivariate canonical adaptation (MCA). The multivariate formulation is initially normalized by IUN for intra-modal unity, and then we optimize the inter-modal (\textit{i.e.} cross-modal data) structure using the constructed space, allowing us to utilize the feature maps learned in the previous step. Many domain-related studies~\cite{huang2017cross,fan2014cross,long2016deep} are based on the concept of ‘feature shift’, which can be summarized as learning a domain-invariant feature representation between the source and target domains for objective transformation. Motivated by the state-of-the-art domain adaptation works~\cite{peng2019moment,gholami2020unsupervised}, we cross-transfer both intra-modal and inter-modal features to a high-level projective space to handle the multivariate heterogeneous medical data. Specifically, we begin by importing a reproducing kernel Hilbert space (RKHS), where the learned features between the CSC-Net layers and RKHS can be formed to minimize the maximum mean discrepancy (MMD)~\cite{gretton2012optimal} between the source and target domains optimally, using a kernel $k(x_i, y_j)= \left \langle \hat{\mathbf{Z}}^x_i,\hat{\mathbf{Z}}^y_j \right \rangle_\mathcal{H}$ under a Hilbert space $\mathcal{H}$. Note that, $\left \langle \cdot,\cdot \right \rangle_\mathcal{H}$ represents the inner product, and $k$ is defined on the vector. Following the virtue of the MMD function in Eq. (\ref{eq2}),
\begin{equation}
\begin{aligned}
\label{eq2}
\mathcal{L}(X,Y)=&\frac{1}{S^2}\sum_{i=1}^{S}\sum_{j=1}^{S}k(\hat{\mathbf{Z}}_i^x,\hat{\mathbf{Z}}_j^x)
+\frac{1}{T^2}\sum_{i=1}^{T}\sum_{j=1}^{T}k^(\hat{\mathbf{Z}}_i^y,\hat{\mathbf{Z}}_j^y)\\
-&\frac{2}{ST}\sum_{i=1}^{S}\sum_{j=1}^{T}k(\hat{\mathbf{Z}}_i^x,\hat{\mathbf{Z}}_j^y),
\end{aligned}
\end{equation}
we adapt the multiple feature layers using the multilayer MMD penalty (termed as $\mathcal{L}_{\mathcal{H}}(X,Y)$) for adapting the cross-modal CSC-Net, which is defined as
\begin{equation}
\begin{aligned}
\label{eq3}
\mathcal{L}_{\mathcal{H}}(X,Y)=&\frac{1}{S^2}\sum_{i=1}^{S}\sum_{j=1}^{S}\prod_{l\in L}k^{\left | l \right |}(\hat{\mathbf{Z}}_i^{x,\left | l \right |},\hat{\mathbf{Z}}_j^{x,\left | l \right |})\\
+&\frac{1}{T^2}\sum_{i=1}^{T}\sum_{j=1}^{T}\prod_{l\in L}k^{\left | l \right |}(\hat{\mathbf{Z}}_i^{y,\left | l \right |},\hat{\mathbf{Z}}_j^{y,\left | l \right |})\\
-&\frac{2}{ST}\sum_{i=1}^{S}\sum_{j=1}^{T}\prod_{l\in L}k^{\left | l \right |}(\hat{\mathbf{Z}}_i^{x,\left | l \right |},\hat{\mathbf{Z}}_j^{y,\left | l \right |}).
\end{aligned}
\end{equation}

Here, the previously described kernel $k$ is computed by the Gaussian kernel function defined on the vectorization of tensors $\hat{\mathbf{Z}}^{x,\left | l \right |}_i \in \mathbb{R}^{N^{\left | l \right |} \times h^{\left | l \right |} \times w^{\left | l \right |}}$ and $\hat{\mathbf{Z}}^{y,\left | l \right |}_j \in \mathbb{R}^{N^{\left | l \right |} \times h^{\left | l \right |} \times w^{\left | l \right |}}$ of the layer $l \in L$. The calculation in a RKHS space can be expressed as $e^{{-\left \| \hat{\mathbf{Z}}_i^{x,\left | l \right |} -\hat{\mathbf{Z}}_j^{y,\left | l \right |} \right \|}^{\frac{2}{p}}}$, where $p$ is a bandwidth parameter reflected in the Gaussian kernel function. $\prod_{l\in L}k^{\left | l \right |}(\hat{\mathbf{Z}}_i^{x,\left | l \right |},\hat{\mathbf{Z}}_j^{y,\left | l \right |})$ gives the multiplicative results of the inner products. 

\subsubsection{Geometric Matching}
While $\mathcal{L}_{\mathcal{H}}(X,Y)$ is endowed with attractive properties for domain invariant representations, its inability to preserve the natural vector structure of a domain leads to the geometric mismatching problem. In addition, deriving an approach that both has an adequate transformation capability and satisfying the neuroimaging constraints is a key challenge. For medical imaging, the positive definite (PD) condition is necessary for diffeomorphisms~\cite{basser1994mr}. This regulation (\textit{i.e.} PD) is omitted in most image synthesis works. As a result, the image being approximated is superficially consistent, but the underlying tissue information or structures are incorrect. To ensure the correctness of medical image synthesis, both information geometry and the PD condition are considered within a Riemannian manifold (RM)~\cite{petersen2006riemannian}. Building on the multilayer MMD introduced earlier, suppose the RM is also a smooth manifold constructed in the Hilbert space $\mathcal{H}$. Theoretically, the RM is equipped with an inner product, denoted as $\left \langle \cdot,\cdot \right \rangle_\mathcal{T}$, on each tangent space $\mathcal{T}$ having manifold $\mathcal{M}$, which can be defined as $\mathcal{T}\mathcal{M}$. As a consequence, the norm in the tangent space is equivalent to that of the Hilbert space $\mathcal{T}\mathcal{M}\cong \mathcal{H}$. 

To make the sparse feature maps respect their intrinsic geometries, we assume that all learned maps comply with the original data properties, \textit{i.e.} the distances between data structures, as well as their corresponding maps, are closed in the RM. Particularly for measurements under high-level feature maps, the main point of RM is to make sense of the image transformation in the manifold setting. Inspired by the benefits of joint learning~\cite{huang2017cross}, we follow such a strategy and model an associated function for converting $\mathbf{Z}^x$ to $\mathbf{Z}^y$. This is done by a linear projection $\mathbf{P}$ of $\mathbf{Z}^x$ and $\mathbf{Z}^y$ using a least square problem: $\left \| \hat{\mathbf{Z}}^y-\mathbf{P}\hat{\mathbf{Z}}^x \right \|_2$. We then recall how the RM can be defined analogously for preserving data fidelity over the image transformation term. We begin by giving the Riemannian metric $\left \langle \hat{\mathbf{Z}}^x_i,\hat{\mathbf{Z}}^y_j \right \rangle_\mathcal{T}$ for $\hat{\mathbf{Z}}^x_i$ and $\hat{\mathbf{Z}}^y_j$ in $\mathcal{T}\mathcal{M}$, and then we rewrite the associated loss in the above least square problem by computing distances on the manifold as $d_\mathcal{M}(\hat{\mathbf{Z}}^y,\mathbf{P}\hat{\mathbf{Z}}^x)=\left \| \log (\hat{\mathbf{Z}}^y)^{-\frac{1}{2}} \mathbf{P}\hat{\mathbf{Z}}^x\log (\hat{\mathbf{Z}}^y)^{-\frac{1}{2}}\right \|_2$, where $\log$ denotes the matrix logarithm and $d_\mathcal{M}$ is affine invariant. Note that $d_\mathcal{M}$ is computed on RA by projecting the symmetric data from $\mathcal{T}$ onto the manifold with the positive definite property~\cite{bhatia2009positive}. For the $l$-layer CSC-Net, the coordinate representation of $d_\mathcal{M}$ is
\begin{equation}
\begin{aligned}
\label{eq4}
\mathcal{L}_\mathcal{M}(X,Y)=\sum_{i=1}^{S}\sum_{j=1}^{T}\prod_{l\in L}d_\mathcal{M}^{\left | l \right |}(\hat{\mathbf{Z}}_i^{y,\left | l \right |},\mathbf{P}_{i,j}^{\left | l \right |}\hat{\mathbf{Z}}_j^{x,\left | l \right |}).
\end{aligned}
\end{equation}

Here, $\mathcal{L}_\mathcal{M}(X,Y)$ denotes the loss function for the multi-layer cross-modal data fidelity of the RM penalty under the positive definite condition.

\subsection{$\ell_4$ Maximization}
Generally, CSC consists of minimizing a convolutional model-fitted least-square system and a sparse regularizer by adopting an $\ell_0$ or $\ell_1$ penalty to promote the expected sparsity for recovering objectives. The $\ell_0$-norm is NP-hard when finding a local minimum of a nonconvex function, while the $\ell_1$-norm provides a unique solution through a convex program in the polynomial time. Although existing CSC algorithms can be separately optimized by alternating subproblems to update sparse feature maps and filters under the non-smooth $\ell_1$ penalty and the $\ell_2$ constraint, when applied to neuroimaging (which has a high-dimensional character), this leads to a high computational complexity due to the quasi-polynomial problem.  

Pioneering research~\cite{zhai2019complete} shows that maximizing the element-wise $\ell_4$-norm enables an interpretative, stable and robust algorithm for reducing the per-iteration cost. Moreover, the global geometry of the $\ell_4$-norm over a unit $\ell_2$-sphere guarantees a randomly initialized first-order gradient descent algorithm~\cite{zhang2019structured}, since each saddle point has negative curvature. Based on the essence of the $\ell_4$-penalized heuristic formulation, \textit{i.e.} replacing $\min \left \| \cdot \right \|_1$ as $\max \left \| \cdot \right \|_4^4$, we attempt to address the shortcomings of CSC by modeling $\ell_4$ maximization directly as the sparsity penalty within the objective of CSC$\ell_4$Net, which can be formulated as
\begin{equation}
\begin{aligned}
\label{eq5}
\min_{\mathcal{L}_\mathcal{M}\mathcal{L}_\mathcal{H},\mathcal{L}_\mathcal{R}}\max_{\mathcal{S}}
&\lambda(\mathcal{S}(X)+\mathcal{S}(Y))+\mathcal{L}_\mathcal{R}(X)+\mathcal{L}_\mathcal{R}(Y)\\
+&\mathcal{L}_\mathcal{M}(X,Y)+\mathcal{L}_{\mathcal{H}}(X,Y).
\end{aligned}
\end{equation}

Here, $\mathcal{S}(X)=\sum_{i=1}^{S}\left \| \hat{\mathbf{Z}}_i^{x,\left | l \right |} \right \|_{4}^4$ and $\mathcal{S}(Y)=\sum_{j=1}^{T}\left \| \hat{\mathbf{Z}}_j^{y,\left | l \right |} \right \|_{4}^4$ represent the sparsity cost function for penalizing $\mathbf{Z}^x$ and $\mathbf{Z}^y$ to be sparse. Mathematically, $\left \| \cdot \right \|_{4}^4$ is the element-wise $\ell_4$-norm with an expression $\left \| \mathbf{Z} \right \|_{4}^4 = \sum_{i,j} \mathbf{Z}_{i,j}^4, \forall \mathbf{Z}\in \mathbb{R}$. $\mathcal{L}_\mathcal{R}(X)$ denotes the reconstruction loss of $\mathbf{X}$ with $\frac{1}{2}\left \| \mathbf{X}-\sum_{i=1}^{S}\mathbf{F}_{i}^{x,\left | l \right |}\ast \hat{\mathbf{Z}}_i^{x,\left | l \right |} \right \|_{2}^{2}$ and $\mathcal{L}_\mathcal{R}(Y)$ is the reconstruction loss of $\mathbf{Y}$ having $\frac{1}{2}\left \| \mathbf{Y}-\sum_{j=1}^{T}\mathbf{F}_{j}^{y,\left | l \right |}\ast \hat{\mathbf{Z}}_j^{y,\left | l \right |} \right \|_{2}^{2}$.

Accompanied with the reconstruction procedure and associated formulation to obtain the target-modality data from the available source domain, we should solve the two respective problems: $\mathbf{Z}$ and $\mathbf{F}$. In this work, to motivate $\ell_4$-based formulations, we employ the matching, stretching and projection (MSP)~\cite{zhai2019complete} optimization method to solve our objective. Simply, the single-layer feature matrix $\mathbf{Z}$ is optimized by introducing MSP on an orthogonal matrix $\mathbf{F}_o \in \mathbb{O}$ with ${\mathbf{Z}^x}'=\arg\max f(\mathbf{X}, \mathbf{\Psi}^x)$, ${\mathbf{Z}^y}'=\arg\max f(\mathbf{Y}, \mathbf{\Psi}^y)$. Tackling the problem of a multi-layer CSC$\ell_4$Net, the $l$-th layer of $\mathbf{Z}$ can be expressed as 
\begin{equation}
\begin{aligned}
\label{eq6}
&\hat{\mathbf{Z}}^{x,\left | l \right |'}_i \leftarrow \arg \max \left \| \hat{\mathbf{Z}}^{x,\left | l \right |}_i \right \|_4^4 \ s.t. \ \mathbf{X}_i^{\left | l \right |} = \mathbf{F}^{x,\left | l-1 \right |}_i \ast \hat{\mathbf{Z}}^{x,\left | l \right |}_i,\\
&\hat{\mathbf{Z}}^{y,\left | l \right |'}_j \leftarrow \arg \max \left \| \hat{\mathbf{Z}}^{y,\left | l \right |}_j \right \|_4^4 \ s.t. \ \mathbf{Y}_j^{\left | l \right |} = \mathbf{F}^{y,\left | l-1 \right |}_j \ast \hat{\mathbf{Z}}^{y,\left | l \right |}_j,
\end{aligned}
\end{equation}
\noindent where $\mathbf{X}_i^{\left | l \right |}=\phi (\hat{\mathbf{Z}}^{x,\left | l \right |}_i)$ and $\mathbf{Y}_j^{\left | l \right |}=\phi (\hat{\mathbf{Z}}^{y,\left | l \right |}_j)$, and $\phi(\cdot)$ denotes the concatenation of all $i$-th features from the previous layer $l-1$. We summarize our CSC$\ell_4$Net in Algorithm \ref{alg1}.
\begin{algorithm}[t!]
	\renewcommand{\algorithmicrequire}{\textbf{Input:}}
	\renewcommand{\algorithmicensure}{\textbf{Output:}}
	\caption{CSC$\ell_4$Net}
	\label{alg1}
	\begin{algorithmic}[1]
		\REQUIRE Training data $\mathbf{X}$, $\mathbf{Y}$, parameter $\lambda$
		\STATE Initialize: $\mathbf{Z}^{x}_0$, $\mathbf{Z}^{y}_0$, $\mathbf{F}^x_0 \in \mathbb{O}$, $\mathbf{F}^y_0 \in \mathbb{O}$, $\mathbf{P}_0$
		\STATE Perform IUN by Eq. (\ref{eq1.1}), $\mathbf{Z}^{x}_0 \rightarrow \hat{\mathbf{Z}}^{x}_0$, $\mathbf{Z}^{y}_0 \rightarrow \hat{\mathbf{Z}}^{y}_0$
		\FOR{$l=1,...L$}
		\STATE $\hat{\mathbf{Z}}^{x,\left | l \right |'}_i \leftarrow \arg \max \left \| \hat{\mathbf{Z}}^{x,\left | l \right |}_i \right \|_4^4 \ s.t. \mathbf{X}_i^{\left | l \right |} = \mathbf{F}^{x,\left | l-1 \right |}_i \ast \hat{\mathbf{Z}}^{x,\left | l \right |}_i$\\
		$\hat{\mathbf{Z}}^{y,\left | l \right |'}_j \leftarrow \arg \max \left \| \hat{\mathbf{Z}}^{y,\left | l \right |}_j \right \|_4^4 \ s.t.  \mathbf{Y}_j^{\left | l \right |} = \mathbf{F}^{y,\left | l-1 \right |}_j \ast \hat{\mathbf{Z}}^{y,\left | l \right |}_j$
		\STATE Perform MSP, update $\mathbf{P}$
		\ENDFOR
		\ENSURE $\mathbf{F}^x$, $\mathbf{F}^y$, $\mathbf{P}$
	\end{algorithmic}  
\end{algorithm}

\subsection{Synthesis}
Once the training stage is complete, we can get two sets of filters, $\mathbf{F}^x$ and $\mathbf{F}^y$, and the associator $\mathbf{P}$. Given a test image $\mathbf{T}^x$ with the source modality, the desirable target modality version of $\mathbf{T}^x$ can be treated as $\mathbf{T}^y=\sum (\mathbf{Z}^{tx}\mathbf{P})\ast \mathbf{F}_{y}$, where $\mathbf{Z}^{tx}=f(\mathbf{T}^x, \mathbf{\Psi}^x)$. 

\section{Experiments}
\subsection{Experimental Setup}\label{5.1}
The proposed CSC$\ell_4$Net is evaluated on three datasets: IXI,~\footnote{http://brain-development.org/ixi-dataset} NAMIC Multimodality,~\footnote{http://insight-journal.org/midas/collection/view/190} and BraTS.~\footnote{https://www.med.upenn.edu/sbia/brats2018/data.html} IXI dataset contains 578 healthy subjects, NAMIC dataset includes 10 normal controls and 10 schizophrenic cases, and BraTS dataset has 220 brain tumor subjects. We conduct extensive experiments on three scenarios: (1) Proton Density (PD) $\rightleftharpoons$ T2 on IXI dataset, (2) T1 $\rightleftharpoons$ T2 on NAMIC dataset, (3) FLAIR $\rightleftharpoons$ T1 on BraTS. In each dataset, the existing well-aligned paired images are removed, with half of them per a side for a strict unsupervised setting. Specifically, 150 unpaired PD-w and T2-w MRI are selected from IXI dataset, 6 unpaired T1-w, T2-w acquisitions are picked from NAMIC dataset, and 70 unpaired T1-w and FLAIR data are chosen from BraTS dataset for training. The split 50 samples from IXI, 2 samples from NAMIC and 30 samples from BraTS are used for validation. The remaining data: 100 (IXI), 4 (NAMIC), and 40 (BraTS) are used for testing. It is worth noting that, in our experimental datasets, IXI only contains healthy images while both NAMIC and BraTS include pathological data. In other words, the first setting in our experiments considers healthy cases, the second setting involves a mix of healthy and pathological examples, and the third setting tests our method on common diseases. We tune the hyper-parameters of our model on the validation set. To verify whether the synthesized results can replace the ground truths with diagnostic acceptability, we feed both real scans and all generations into a classic and commonly used segmentation algorithm, FMRIB software library (FSL)~\cite{jenkinson2012fsl}. We classify cerebrospinal fluid (CSF), gray matter (GM), white matter (WM) and optional tumor lesions to obtain the average quantification of the whole brain volume. FSL\footnote{https://fsl.fmrib.ox.ac.uk/fsl/fslwiki} is a comprehensive library of analysis, specific for functional and structural brain imaging data. We obtain the segmented results under the tissue prior probability templates in a default image space, and therefore there is no guarantee that our segmentation will exactly follow other methods. For evaluation criteria, we use PSNR, SSIM and Dice score (which measures segmentation overlap, with a higher value meaning a better result) to assess the performance of different methods. We demonstrate that the proposed CSC$\ell_4$Net exhibits competitive performance with fewer layers (and the corresponding parameters) compared to other deep learning methods.
\begin{figure}[t!]
	\centering
	\includegraphics[width=1\linewidth]{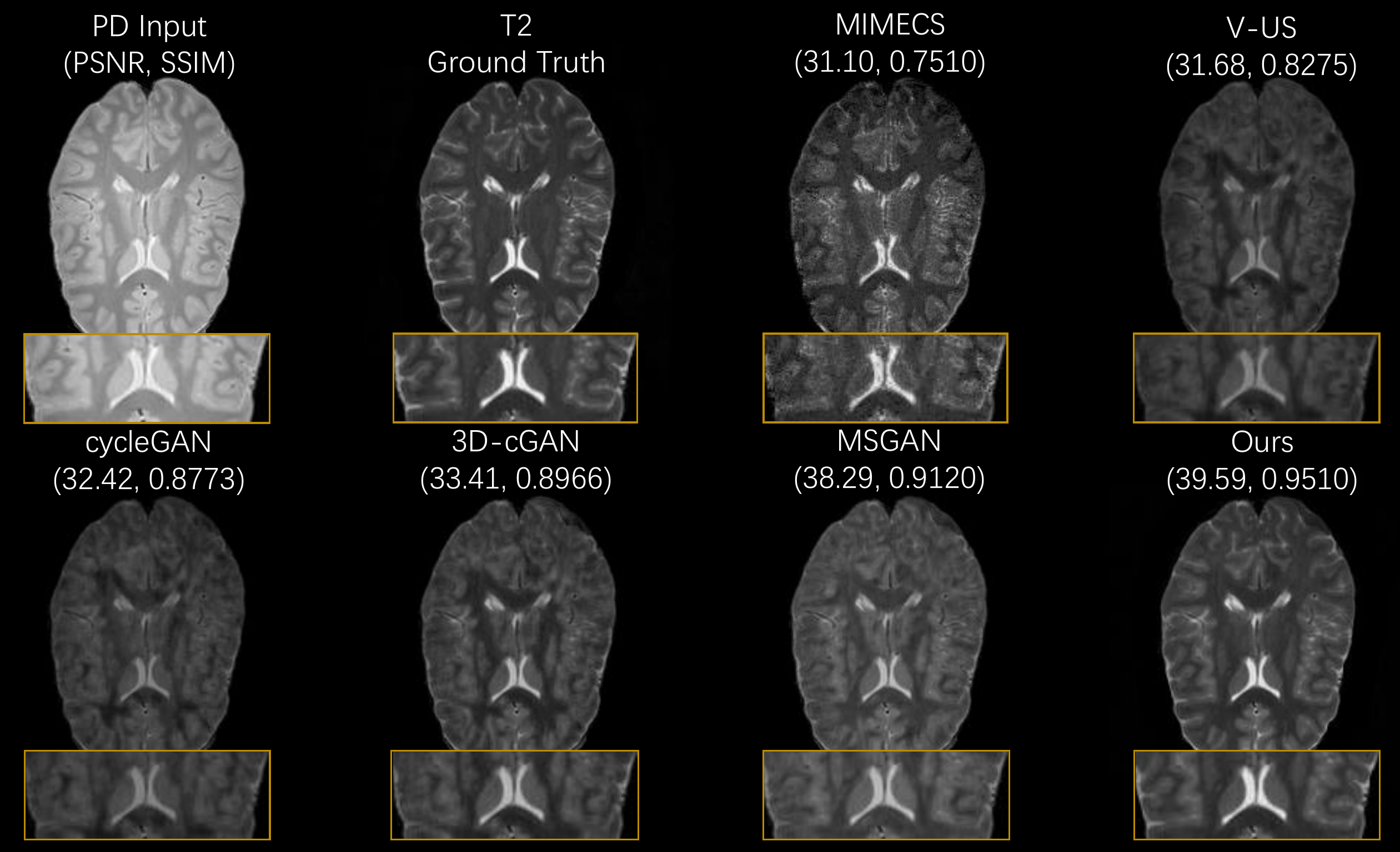}
	\caption{Visual comparisons of MIMECS, V-US, cycleGAN, 3D-cGAN, MSGAN and ours for PD$\rightarrow$T2 on IXI dataset.}
	\label{fig1}
\end{figure}

\subsection{Network Architecture}
Inspired by ResNet~\cite{he2016deep}, CSC$\ell_4$Net includes seven CSC layers with layered MMD and RM constraints, where the spatial subsampling operation is performed with a stride of 2 in the last two bottleneck layers. Batch normalization is incorporated layer-wise, following each CSC layer, for faster convergence. The last CSC layer lies on top of a global spatial average pooling layer. The bottleneck layer in CSC$\ell_4$Net outputs 64, 128, 256 with 3, 2, 2 times repeated stacks, respectively. We use a learning rate of 0.0002, a batch size of 16, a total of 200 epochs, and a sparse regularization parameter of 0.02. We work in 2D slices and employ the Adam optimizer. For the layered MMD parameters, we use the Gaussian kernel with bandwidths equipped as median pairwise squared distances. We use the same network configurations for all datasets.

\subsection{Baseline Methods}
We compare our CSC$\ell_4$Net with the most relevant and state-of-the-art synthesis methods: 1) MIMECS \cite{roy2013magnetic}; 2) V-US \cite{vemulapalli2015unsupervised}; 3) cycleGAN \cite{zhu2017unpaired}; 4) 3D-cGAN \cite{pan2018synthesizing}; 5) MSGAN \cite{mao2019mode}. Note that, MIMECS and MSGAN are supervised methods, so we input originally paired data for their training. V-US, cycleGAN and 3D-cGAN are unsupervised approaches, so we input the selected unpaired images for training. For fair comparison, we empirically set all parameters of the compared methods following their recommended data to reach their best performance. Notably, we exactly followed the data processing steps implemented by MIMECS, V-US and 3D-cGAN, where a standard intensity correction and skull-striping were preformed. Therefore, no extra data processes are implemented.

\begin{figure}[t!]
	\centering
	\includegraphics[width=1\linewidth]{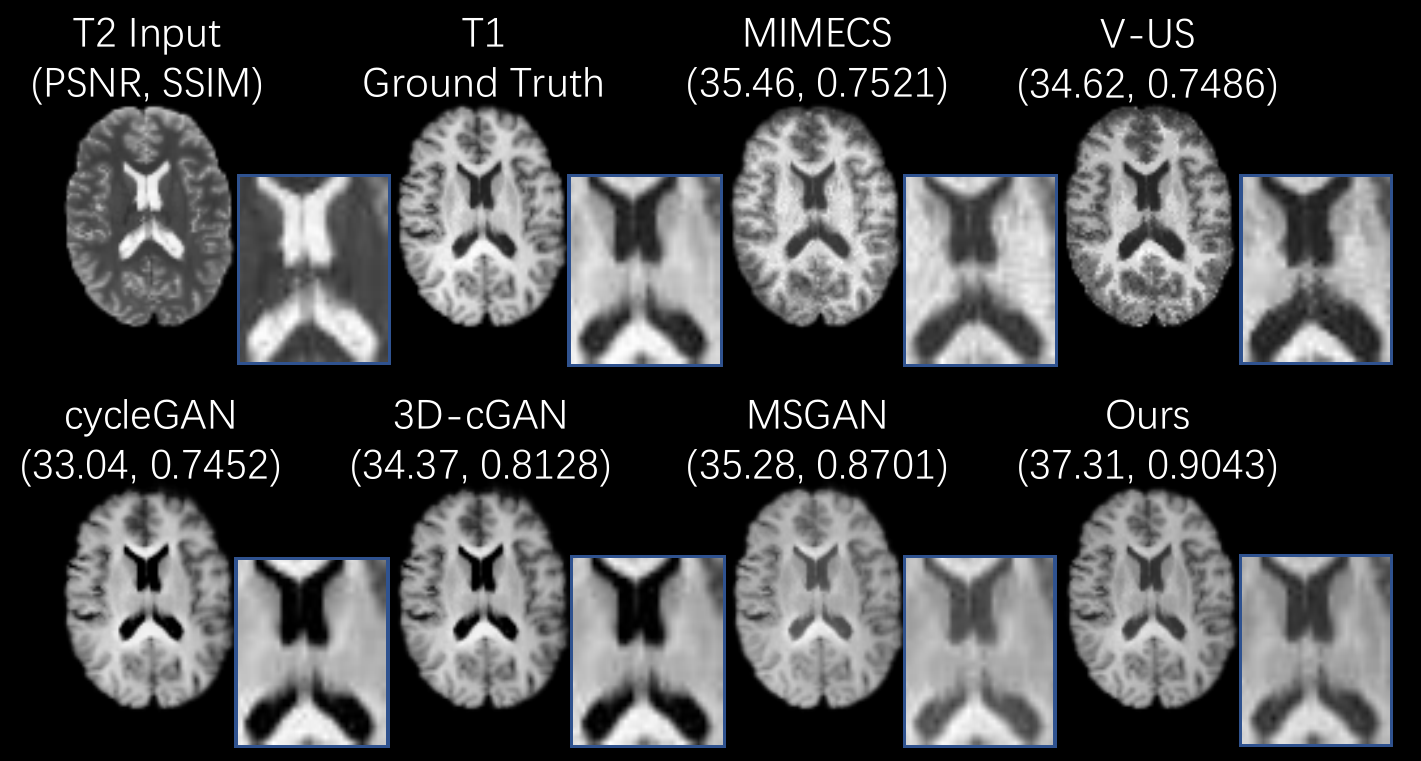}
	\caption{Visual comparisons of MIMECS, V-US, cycleGAN, 3D-cGAN, MSGAN and ours for T2$\rightarrow$T1 on NAMIC dataset.}
	\label{fig2}
\end{figure}
\begin{table}[t!]
	\begin{center}
		\scalebox{0.68}{
		\begin{tabular}{c|c c c c c c}
			\hline
			Metric(avg.) & MIMECS & V-US & cycleGAN & 3D-cGAN & MSGAN & CSC$\ell_4$Net \\
			\hline
			\hline
			\multicolumn{7}{c}{IXI: T2-w $\rightarrow$ PD-w}\\
			\hline
			PSNR (dB) & \textit{30.49} & 32.53 & 32.66 & 32.75 & 33.11 & \textbf{36.45} \\
			SSIM & \textit{0.7801} & 0.8322 & 0.8345 & 0.8517 & 0.8557 & \textbf{0.9065} \\
			Dice (in \%) & 72.59 & \textit{67.73} & 68.89 & 76.87 & 72.92 & \textbf{80.78} \\
			\hline
			\multicolumn{7}{c}{IXI: PD-w $\rightarrow$ T2-w}\\
			\hline
			PSNR (dB) & \textit{31.64} & 33.94 & 34.34 & 34.97 & 35.29 & \textbf{38.18}\\
			SSIM & \textit{0.8170} & 0.9022 & 0.9085 & 0.9035 & 0.8971 & \textbf{0.9596}\\
			Dice (in \%) & 76.52 & \textit{69.87} & 70.26 & 80.63 & 79.78 & \textbf{87.90} \\
			\hline
			\hline
			\multicolumn{7}{c}{NAMIC: T1-w $\rightarrow$ T2-w}\\
			\hline
			PSNR (dB) & 35.97 & 35.50 & 36.41 & \textit{35.44} & 35.78 & \textbf{37.82} \\
			SSIM & 0.7690 & \textit{0.7638} & 0.8291 & 0.8846 & 0.9033 & \textbf{0.9613} \\
			Dice (in \%) & 70.47 & 69.98 & \textit{69.23} & 71.67 & 71.33 & \textbf{85.21} \\
			\hline
			\multicolumn{7}{c}{NAMIC: T2-w $\rightarrow$ T1-w}\\
			\hline
			PSNR (dB) & 34.50 & 34.38 & 34.43 & \textit{34.34} & 35.26 & \textbf{37.00} \\
			SSIM & \textit{0.7389} & 0.7517 & 0.8135 & 0.8673 & 0.8883 & \textbf{0.9222} \\
			Dice (in \%) & 72.19 & 70.09 & \textit{70.07} & 74.62 & 72.69 & \textbf{83.64}\\
			\hline
			\hline
			\multicolumn{7}{c}{BraTS: T1-w $\rightarrow$ FLAIR}\\
			\hline
			PSNR (dB) & \textit{30.50} & 31.87 & 31.37 & 33.74 & 31.81 & \textbf{37.41}\\
			SSIM & \textit{0.7944} & 0.8341 & 0.8129 & 0.8763 & 0.8653 & \textbf{0.9344}\\
			Dice (in \%) & 70.55 & \textit{69.23} & 71.96 & 73.87 & 73.92 & \textbf{83.98}\\
			\hline
			\multicolumn{7}{c}{BraTS: FLAIR $\rightarrow$ T1-w}\\
			\hline
			PSNR (dB) & \textit{30.04} & 31.81 & 31.25 & 32.87 & 33.69 & \textbf{36.46}\\
			SSIM & \textit{0.8125} & 0.8421 & 0.8495 & 0.8798 & 0.8607 & \textbf{0.9112}\\
			Dice (in \%) & 71.59 & \textit{69.83} & 73.21 & 78.87 & 76.99 & \textbf{82.63} \\
			\hline
		\end{tabular}}
	\end{center}
	\caption{The performance of synthesis and synthesis-based segmentation on IXI, NAMIC and BraTS datasets.}
	\label{tab1}
\end{table}
\begin{figure}[t!]
	\centering
	\includegraphics[width=1\linewidth]{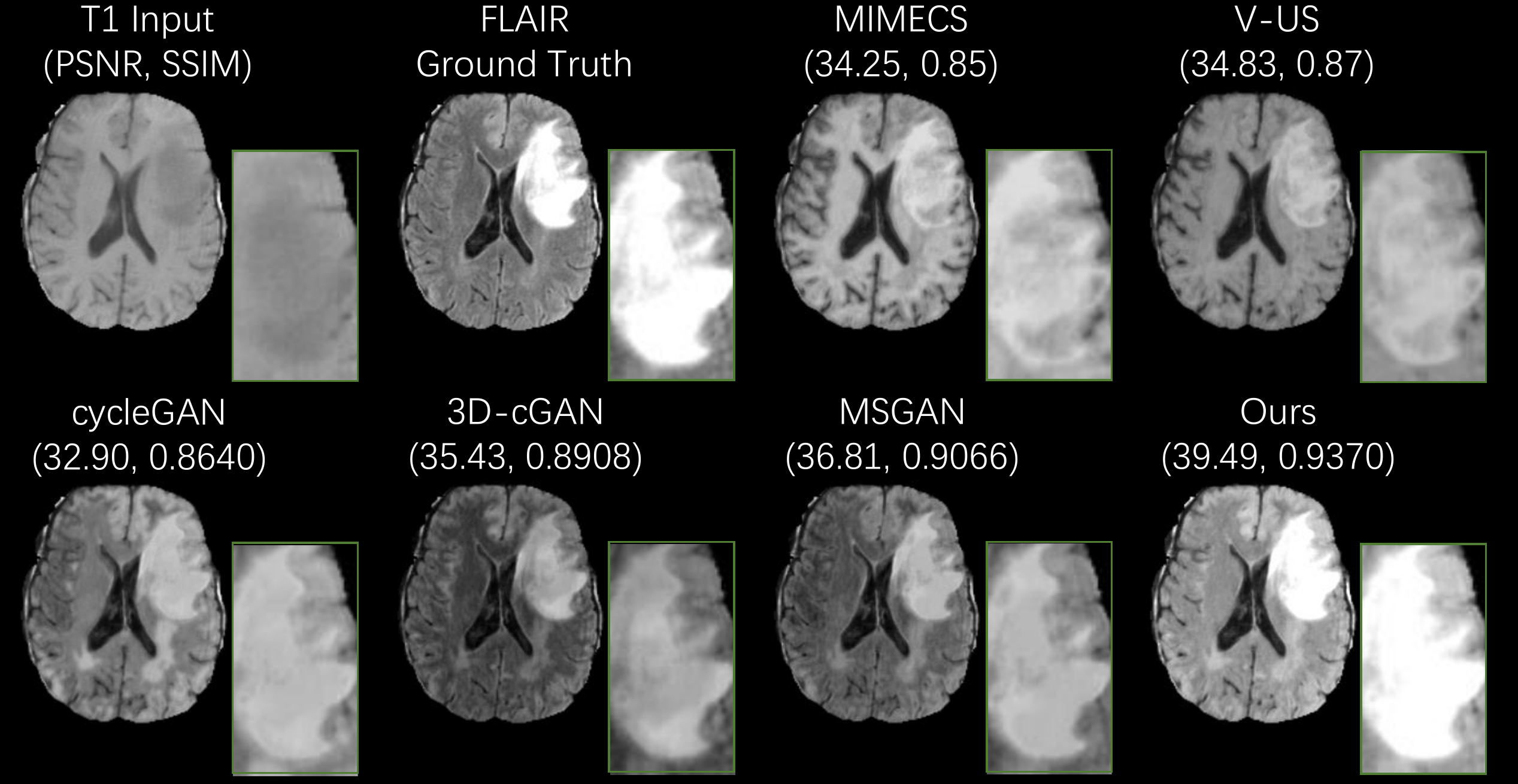}
	\caption{Visual comparisons of MIMECS, V-US, cycleGAN, 3D-cGAN, MSGAN and ours for T1$\rightarrow$FLAIR on BraTS dataset.}
	\label{fig3}
	\vspace{-0.2cm}
\end{figure}

\subsection{Results}
We evaluate the effectiveness of our method by synthesizing images on different datasets. The results of CSC$\ell_4$Net and the five baseline methods are compared in Table~\ref{tab1}. As further validation, we also apply the synthesized results to segment major tissues in each dataset and summarize the average performances in Table~\ref{tab1}. First, we compare our CSC$\ell_4$Net against several state-of-the-art synthesis methods on IXI for transforming PD-w MRI to T2-w MRI and \textit{vice versa}. For better interpretation, we provide the visualization results in Fig.~\ref{fig1}. From both Table \ref{tab1} (top) and Fig.~\ref{fig1}, we observe that CSC$\ell_4$Net achieves much better performance (visually and quantitatively) than the other five baseline methods in all cases. The average synthesis performances of CSC$\ell_4$Net on IXI dataset are \{PSNR: 36.45dB, SSIM: 0.9065\} and \{PSNR: 38.18dB, SSIM: 0.9596\} for T2-w $\rightarrow$ PD-w and PD-w $\rightarrow$ T2-w, respectively. CSC$\ell_4$Net gains a significant \{PSNR, SSIM\} performance improvement of \{3.34dB, 0.0508\} and \{2.89dB, 0.0625\}, respectively, compared to the best baseline MSGAN. In addition, we report the segmentation accuracy (\textit{Dice scores}), which are based on the synthesized results, in Table \ref{tab1}. As we intuitively expected, if the model can provide results with data fidelity, the synthesis task is able to deliver practical usable results for further diagnosis. For medical image synthesis-driven segmentation, our method again outperforms the other methods, \textit{i.e.,} MIMECS, V-US, cycleGAN, 3D-cGAN and MSGAN, by a large margin. Our results are more representative than those of other state-of-the-art image synthesis methods on IXI dataset, showing dice improvements of 3.91\% and 7.27\% for T2-w $\rightarrow$ PD-w and PD-w $\rightarrow$ T2-w respectively, \textit{w.r.t} the best baseline.
\begin{figure*}[t!]
	\centering
	\includegraphics[width=1\linewidth]{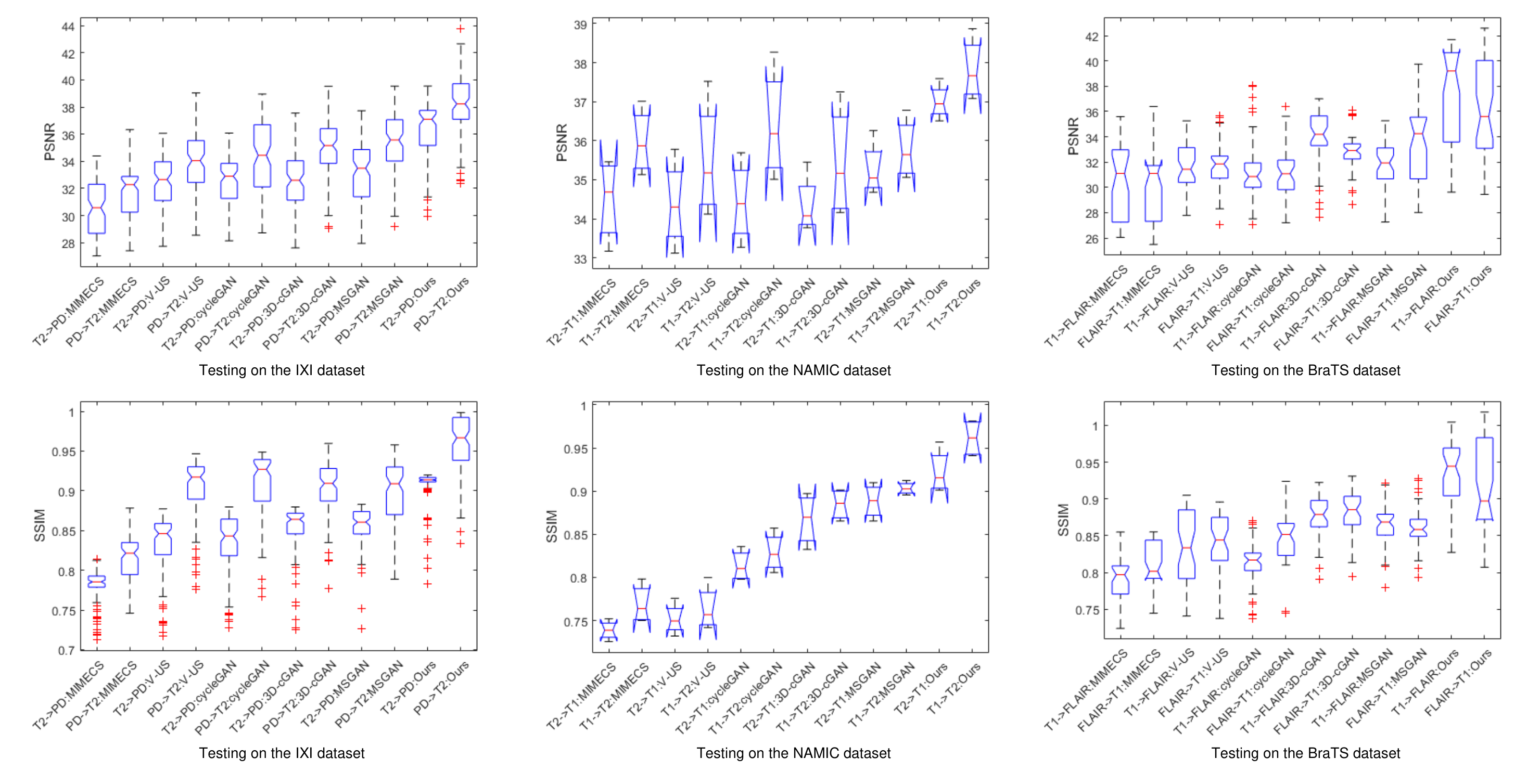}
	\caption{Quantitative evaluations including PSNRs and SSIMs of all methods distributed across IXI, NAMIC and BraTS, respectively.}
	\label{fig4}
\end{figure*}

In addition to healthy cases, we also evaluate our algorithm on the pathological datasets (\textit{i.e.} NAMIC, BraTS). We adopt the same evaluation criteria as on IXI dataset. We provide the qualitative results of our method in Figs.~\ref{fig2}-\ref{fig3}. These figures demonstrate how BrainGAN can handle schizophrenic and brain tumor synthesis on NAMIC and BraTS dataset, respectively. The proposed method consistently obtains the best performance, particularly when the lesions in the T1-w data appear with much lower contrast than in the FLAIR brain MRI (shown in Fig.~\ref{fig3}). In Table~\ref{tab1} (middle and bottom), similarly for PD $\rightleftharpoons$ T2 synthesis, CSC$\ell_4$Net outperforms the compared methods with improvements of \{1.41dB, 0.0580\} and \{1.74dB, 0.0339\} for T1 $\rightleftharpoons$ T2 on NAMIC dataset, and \{3.67dB, 0.0582\} and \{2.77dB, 0.0314\} for T1 $\rightleftharpoons$ FLAIR on BraTS dataset, respectively. Using the defined settings (refer to Section~\ref{5.1}) and all synthesized data from both the forward and backward transformation, Table~\ref{tab1} compares the segmentation results. As with the Dice scores reported in Table \ref{tab1} (middle and bottom), our method again gains significant performance enhancements, \textit{i.e.}, increasing by 13.54\% for T1 $\rightleftharpoons$ T2, 9.02\% for T2 $\rightleftharpoons$ T1, 9.72\% for T1 $\rightleftharpoons$ FLAIR, and 3.76\% for FLAIR $\rightleftharpoons$ T1. For clarity, Fig.~\ref{fig4} provides detailed comparison results of all methods distributed across different datasets.

\subsection{Ablation Studies}
Since CSC$\ell_4$Net comprises a combination of several components, we perform an extensive ablation study to better understand why it is able to obtain state-of-the-art results. Considering the number of experiments in the above studies, we focus on the case of PD-w $\rightarrow$ T2-w from IXI dataset and report our ablation results in Table~\ref{tab2}. We separately adopt \{IUN, $\mathcal{L}_\mathcal{H}$, $\mathcal{L}_\mathcal{M}$\} as the additional module under the baseline CSC, and evaluate the effects in terms of image quality and synthesis-based segmentation performance. When each of the components is separately combined with CSC, the PSNR, SSIM, and Dice scores are improved by \{2.42dB, 3.73dB, 3.54dB\}, \{0.05, 0.07, 0.07\}, \{0.35\%, 2.19\%, 7.79\%\}, respectively. With the assistance of $\mathcal{L}_\mathcal{H}$ and $\mathcal{L}_\mathcal{M}$, the image quality and synthesis-based segmentation have the greatest impact. CSC with different combinations of two modules achieves \{7.42dB. 7.05dB, 9.33dB\}, \{0.16, 0.16, 0.19\}, and \{13.44\%, 17.7\%, 26.43\%\} improvements over the baseline.

\begin{table}[t!]
	\vspace{-0.5cm}
	\begin{center}
		\scalebox{0.78}{
			\begin{tabular}{c c c c|c c c}
				\hline
				CSC & IUN & $\mathcal{L}_\mathcal{H}$ & $\mathcal{L}_\mathcal{M}$ & PSNR(dB) & SSIM & Dice(\%) \\
				\hline
				\hline
				$\checkmark$ & & & & 27.54 & 0.7357 & 58.73\\
				$\checkmark$ & $\checkmark$ & & & 29.96 & 0.7864 & 59.08\\
				$\checkmark$ & & $\checkmark$ & & 31.27 & 0.8022 & 60.92\\
				$\checkmark$ & & & $\checkmark$ & 31.08 & 0.8070 & 66.52\\
				$\checkmark$ & $\checkmark$ & $\checkmark$ & & 34.96 & 0.8958 & 72.17\\
				$\checkmark$ & $\checkmark$ & & $\checkmark$ & 34.59 & 0.8996 & 76.43\\
				$\checkmark$ & & $\checkmark$ & $\checkmark$ & 36.87 & 0.9214 & 85.16\\
				\hline
		\end{tabular}}
	\end{center}
	\caption{Ablation study showing the effect of our modules/loss functions in improving the performance for PD-w $\rightarrow$ T2-w on IXI.}
	\vspace{-0.53cm}
	\label{tab2}
\end{table}

\section{Conclusion}
We have proposed a novel multivariate canonical CSC$\ell_4$Net approach for the cross-modal synthesis of medical images. CSC$\ell_4$Net aims at unifying multivariate datasets across both intra-modal and inter-modal through layer-wise feature adaptation and manifold transformation. CSC$\ell_4$Net is robust against both appearance variation and irregular machines. In addition, the proposed method extends the general CSC to a multi-layer CSC and imposes multivariate canonical feature mapping under the $\ell_4$-maximization to account for the high-dimensional and heterogeneous nature of neuroimaging. Comprehensive experiments show that CSC$\ell_4$Net is effective for a variety of cross-modality medical image synthesis problems, with segmentation quality, and can significantly outperform state-of-the-art methods even for very difference datasets. In the future, we plan to extend our CSC$\ell_4$Net to various image formats to investigate its generality.

\noindent\textbf{Acknowledgment:} This work is supported by the National Natural Science Foundation of China under Grant 61972188.


{\small
\bibliographystyle{ieee_fullname}
\bibliography{egbib}

\begin{thebibliography}{10}\itemsep=-1pt

\bibitem{basser1994mr}
Peter~J Basser, James Mattiello, and Denis LeBihan.
\newblock Mr diffusion tensor spectroscopy and imaging.
\newblock {\em Biophysical journal}, 66(1):259--267, 1994.

\bibitem{bhatia2009positive}
Rajendra Bhatia.
\newblock {\em Positive definite matrices}, volume~24.
\newblock Princeton university press, 2009.

\bibitem{bristow2013fast}
Hilton Bristow, Anders Eriksson, and Simon Lucey.
\newblock Fast convolutional sparse coding.
\newblock In {\em Proceedings of the IEEE Conference on Computer Vision and
  Pattern Recognition}, pages 391--398, 2013.

\bibitem{gatys2016image}
Leon~A Gatys, Alexander~S Ecker, and Matthias Bethge.
\newblock Image style transfer using convolutional neural networks.
\newblock In {\em Proceedings of the IEEE conference on computer vision and
  pattern recognition}, pages 2414--2423, 2016.

\bibitem{gholami2020unsupervised}
Behnam Gholami, Pritish Sahu, Ognjen Rudovic, Konstantinos Bousmalis, and
  Vladimir Pavlovic.
\newblock Unsupervised multi-target domain adaptation: An information theoretic
  approach.
\newblock {\em IEEE Transactions on Image Processing}, 2020.

\bibitem{goodfellow2014generative}
Ian Goodfellow, Jean Pouget-Abadie, Mehdi Mirza, Bing Xu, David Warde-Farley,
  Sherjil Ozair, Aaron Courville, and Yoshua Bengio.
\newblock Generative adversarial nets.
\newblock In {\em Advances in neural information processing systems}, pages
  2672--2680, 2014.

\bibitem{gretton2012optimal}
Arthur Gretton, Dino Sejdinovic, Heiko Strathmann, Sivaraman Balakrishnan,
  Massimiliano Pontil, Kenji Fukumizu, and Bharath~K Sriperumbudur.
\newblock Optimal kernel choice for large-scale two-sample tests.
\newblock In {\em Advances in neural information processing systems}, pages
  1205--1213, 2012.

\bibitem{gu2015convolutional}
Shuhang Gu, Wangmeng Zuo, Qi Xie, Deyu Meng, Xiangchu Feng, and Lei Zhang.
\newblock Convolutional sparse coding for image super-resolution.
\newblock In {\em Proceedings of the IEEE International Conference on Computer
  Vision}, pages 1823--1831, 2015.

\bibitem{he2016deep}
Kaiming He, Xiangyu Zhang, Shaoqing Ren, and Jian Sun.
\newblock Deep residual learning for image recognition.
\newblock In {\em Proceedings of the IEEE conference on computer vision and
  pattern recognition}, pages 770--778, 2016.

\bibitem{heide2015fast}
Felix Heide, Wolfgang Heidrich, and Gordon Wetzstein.
\newblock Fast and flexible convolutional sparse coding.
\newblock In {\em Proceedings of the IEEE Conference on Computer Vision and
  Pattern Recognition}, pages 5135--5143, 2015.

\bibitem{huang2017cross}
Yawen Huang, Ling Shao, and Alejandro~F Frangi.
\newblock Cross-modality image synthesis via weakly coupled and geometry
  co-regularized joint dictionary learning.
\newblock {\em IEEE transactions on medical imaging}, 37(3):815--827, 2017.

\bibitem{huang2017dote}
Yawen Huang, Ling Shao, and Alejandro~F Frangi.
\newblock Dote: Dual convolutional filter learning for super-resolution and
  cross-modality synthesis in mri.
\newblock In {\em International Conference on Medical Image Computing and
  Computer-Assisted Intervention}, pages 89--98. Springer, 2017.

\bibitem{huang2020mcmt}
Yawen Huang, Feng Zheng, Runmin Cong, Weilin Huang, Matthew~R Scott, and Ling
  Shao.
\newblock Mcmt-gan: Multi-task coherent modality transferable gan for 3d brain
  image synthesis.
\newblock {\em IEEE Transactions on Image Processing}, 29:8187--8198, 2020.

\bibitem{iglesias2018joint}
Juan~Eugenio Iglesias, Marc Modat, Lo{\"\i}c Peter, Allison Stevens, Roberto
  Annunziata, Tom Vercauteren, Ed Lein, Bruce Fischl, Sebastien Ourselin,
  Alzheimer’s Disease~Neuroimaging Initiative, et~al.
\newblock Joint registration and synthesis using a probabilistic model for
  alignment of mri and histological sections.
\newblock {\em Medical image analysis}, 50:127--144, 2018.

\bibitem{jenkinson2012fsl}
Mark Jenkinson, Christian~F Beckmann, Timothy~EJ Behrens, Mark~W Woolrich, and
  Stephen~M Smith.
\newblock Fsl.
\newblock {\em Neuroimage}, 62(2):782--790, 2012.

\bibitem{liu2019few}
Ming-Yu Liu, Xun Huang, Arun Mallya, Tero Karras, Timo Aila, Jaakko Lehtinen,
  and Jan Kautz.
\newblock Few-shot unsupervised image-to-image translation.
\newblock {\em arXiv preprint arXiv:1905.01723}, 2019.

\bibitem{long2016deep}
Mingsheng Long, Jianmin Wang, Yue Cao, Jiaguang Sun, and S~Yu Philip.
\newblock Deep learning of transferable representation for scalable domain
  adaptation.
\newblock {\em IEEE Transactions on Knowledge and Data Engineering},
  28(8):2027--2040, 2016.

\bibitem{mao2019mode}
Qi Mao, Hsin-Ying Lee, Hung-Yu Tseng, Siwei Ma, and Ming-Hsuan Yang.
\newblock Mode seeking generative adversarial networks for diverse image
  synthesis.
\newblock In {\em Proceedings of the IEEE Conference on Computer Vision and
  Pattern Recognition}, pages 1429--1437, 2019.

\bibitem{moreau2018dicod}
Thomas Moreau, Laurent Oudre, and Nicolas Vayatis.
\newblock Dicod: Distributed convolutional coordinate descent for convolutional
  sparse coding.
\newblock In {\em International Conference on Machine Learning}, pages
  3623--3631, 2018.

\bibitem{pan2018synthesizing}
Yongsheng Pan, Mingxia Liu, Chunfeng Lian, Tao Zhou, Yong Xia, and Dinggang
  Shen.
\newblock Synthesizing missing pet from mri with cycle-consistent generative
  adversarial networks for alzheimer’s disease diagnosis.
\newblock In {\em MICCAI}, pages 455--463. Springer, 2018.

\bibitem{peng2019moment}
Xingchao Peng, Qinxun Bai, Xide Xia, Zijun Huang, Kate Saenko, and Bo Wang.
\newblock Moment matching for multi-source domain adaptation.
\newblock In {\em Proceedings of the IEEE International Conference on Computer
  Vision}, pages 1406--1415, 2019.

\bibitem{petersen2006riemannian}
Peter Petersen, S Axler, and KA Ribet.
\newblock {\em Riemannian geometry}, volume 171.
\newblock Springer, 2006.

\bibitem{ravi2019degenerative}
Daniele Ravi, Daniel~C Alexander, Neil~P Oxtoby, Alzheimer’s
  Disease~Neuroimaging Initiative, et~al.
\newblock Degenerative adversarial neuroimage nets: Generating images that
  mimic disease progression.
\newblock In {\em International Conference on Medical Image Computing and
  Computer-Assisted Intervention}, pages 164--172. Springer, 2019.

\bibitem{roy2013magnetic}
Snehashis Roy, Aaron Carass, and Jerry~L Prince.
\newblock Magnetic resonance image example-based contrast synthesis.
\newblock {\em IEEE transactions on medical imaging}, 32(12):2348--2363, 2013.

\bibitem{shao2019diagnosis}
Wei Shao, Tongxin Wang, Zhi Huang, Jun Cheng, Zhi Han, Daoqiang Zhang, and Kun
  Huang.
\newblock Diagnosis-guided multi-modal feature selection for prognosis
  prediction of lung squamous cell carcinoma.
\newblock In {\em International Conference on Medical Image Computing and
  Computer-Assisted Intervention}, pages 113--121. Springer, 2019.

\bibitem{sun2019adversarial}
Bo Sun, Nian-hsuan Tsai, Fangchen Liu, Ronald Yu, and Hao Su.
\newblock Adversarial defense by stratified convolutional sparse coding.
\newblock In {\em Proceedings of the IEEE Conference on Computer Vision and
  Pattern Recognition}, pages 11447--11456, 2019.

\bibitem{uzunova2019multi}
Hristina Uzunova, Jan Ehrhardt, Fabian Jacob, Alex Frydrychowicz, and Heinz
  Handels.
\newblock Multi-scale gans for memory-efficient generation of high resolution
  medical images.
\newblock In {\em International Conference on Medical Image Computing and
  Computer-Assisted Intervention}, pages 112--120. Springer, 2019.

\bibitem{vemulapalli2015unsupervised}
Raviteja Vemulapalli, Hien Van~Nguyen, and Shaohua Kevin~Zhou.
\newblock Unsupervised cross-modal synthesis of subject-specific scans.
\newblock In {\em Proceedings of the IEEE International Conference on Computer
  Vision}, pages 630--638, 2015.

\bibitem{wang2020transformation}
Jiayu Wang, Wengang Zhou, Guo-Jun Qi, Zhongqian Fu, Qi Tian, and Houqiang Li.
\newblock Transformation gan for unsupervised image synthesis and
  representation learning.
\newblock In {\em Proceedings of the IEEE/CVF Conference on Computer Vision and
  Pattern Recognition}, pages 472--481, 2020.

\bibitem{wang2018high}
Ting-Chun Wang, Ming-Yu Liu, Jun-Yan Zhu, Andrew Tao, Jan Kautz, and Bryan
  Catanzaro.
\newblock High-resolution image synthesis and semantic manipulation with
  conditional gans.
\newblock In {\em Proceedings of the IEEE conference on computer vision and
  pattern recognition}, pages 8798--8807, 2018.

\bibitem{zeiler2010deconvolutional}
Matthew~D Zeiler, Dilip Krishnan, Graham~W Taylor, and Rob Fergus.
\newblock Deconvolutional networks.
\newblock In {\em 2010 IEEE Computer Society Conference on computer vision and
  pattern recognition}, pages 2528--2535. IEEE, 2010.

\bibitem{zhai2019complete}
Yuexiang Zhai, Zitong Yang, Zhenyu Liao, John Wright, and Yi Ma.
\newblock Complete dictionary learning via $\ell_4$-norm maximization over the
  orthogonal group.
\newblock {\em arXiv preprint arXiv:1906.02435}, 2019.

\bibitem{zhang2019structured}
Yuqian Zhang, Han-Wen Kuo, and John Wright.
\newblock Structured local optima in sparse blind deconvolution.
\newblock {\em IEEE Transactions on Information Theory}, 66(1):419--452, 2019.

\bibitem{zhang2018translating}
Zizhao Zhang, Lin Yang, and Yefeng Zheng.
\newblock Translating and segmenting multimodal medical volumes with cycle-and
  shape-consistency generative adversarial network.
\newblock In {\em Proceedings of the IEEE Conference on Computer Vision and
  Pattern Recognition}, pages 9242--9251, 2018.

\bibitem{zhao2019data}
Amy Zhao, Guha Balakrishnan, Fredo Durand, John~V Guttag, and Adrian~V Dalca.
\newblock Data augmentation using learned transformations for one-shot medical
  image segmentation.
\newblock In {\em Proceedings of the IEEE Conference on Computer Vision and
  Pattern Recognition}, pages 8543--8553, 2019.

\bibitem{feng2019Pyramidal}
Feng Zheng, Cheng Deng, Xing Sun, Xinyang Jiang, Xiaowei Guo, Zongqiao Yu,
  Feiyue Huang, and Rongrong Ji.
\newblock Pyramidal person re-identification via multi-loss dynamic training.
\newblock In {\em Proceedings of the IEEE Conference on Computer Vision and
  Pattern Recognition}, pages 8514--8522, 2019.

\bibitem{zhou2019high}
Yi Zhou, Xiaodong He, Shanshan Cui, Fan Zhu, Li Liu, and Ling Shao.
\newblock High-resolution diabetic retinopathy image synthesis manipulated by
  grading and lesions.
\newblock In {\em International Conference on Medical Image Computing and
  Computer-Assisted Intervention}, pages 505--513. Springer, 2019.

\bibitem{fan2014cross}
Fan Zhu and Ling Shao.
\newblock Weakly-supervised cross-domain dictionary learning for visual
  recognition.
\newblock {\em International Journal of Computer Vision}, 109(1-2):42--59,
  2014.

\bibitem{zhu2017unpaired}
Jun-Yan Zhu, Taesung Park, Phillip Isola, and Alexei~A Efros.
\newblock Unpaired image-to-image translation using cycle-consistent
  adversarial networks.
\newblock In {\em Proceedings of the IEEE international conference on computer
  vision}, pages 2223--2232, 2017.

\bibitem{zisselman2019local}
Ev Zisselman, Jeremias Sulam, and Michael Elad.
\newblock A local block coordinate descent algorithm for the csc model.
\newblock In {\em Proceedings of the IEEE Conference on Computer Vision and
  Pattern Recognition}, pages 8208--8217, 2019.

\end{thebibliography}
}

\end{document}